# GeReA: Question-Aware Prompt Captions for Knowledge-based Visual Question Answering

Ziyu Ma, Shutao Li, *Fellow, IEEE*, Bin Sun, *Member, IEEE*, Jianfei Cai, *Fellow, IEEE*, Zuxiang Long, and Fuyan Ma

**Abstract**—Knowledge-based visual question answering (VQA) requires world knowledge beyond the image for accurate answer. Recently, instead of extra knowledge bases, a large language model (LLM) like GPT-3 [1] is activated as an implicit knowledge engine to jointly acquire and reason the necessary knowledge for answering by converting images into textual information (e.g., captions and answer candidates). However, such conversion may introduce irrelevant information, which causes the LLM to misinterpret images and ignore visual details crucial for accurate knowledge. We argue that multimodal large language model (MLLM) is a better implicit knowledge engine than the LLM for its superior capability of visual understanding. Despite this, how to activate the capacity of MLLM as the implicit knowledge engine has not been explored yet. Therefore, we propose GeReA, a generate-reason framework that prompts a MLLM like InstructBLIP [2] with question relevant vision and language information to generate knowledge-relevant descriptions and reasons those descriptions for knowledge-based VQA. Specifically, the question-relevant image regions and question-specific manual prompts are encoded in the MLLM to generate the knowledge relevant descriptions, referred to as question-aware prompt captions. After that, the question-aware prompt captions, image-question pair, and similar samples are sent into the multi-modal reasoning model to learn a joint knowledge-image-question representation for answer prediction. GeReA unlocks the use of MLLM as the implicit knowledge engine, surpassing all previous state-of-the-art methods on OK-VQA and A-OKVQA datasets, with test accuracies of 66.5% and 63.3% respectively. Our code will be released at https://github.com/Upper9527/GeReA.

✦

## 1 INTRODUCTION

THE visual question answering (VQA) task requires a machine to answer open-domain questions given the image [3]. This task has been extensively explored in the past few years and many multi-modal methods have been proposed and gained state-of-the-art performance [4], [5], [6], [7], [8], [9], [10], [11], [12]. The knowledge-based VQA task [13], [14], [15], [16], [17] extends the traditional VQA task that requires external knowledge beyond the image for accurate answer and necessitates a more comprehensive learning approach. Early knowledge-based VQA benchmarks introduce pre-defined knowledge bases (KBs), which are included in the majority of knowledge-based VQA datasets, and each question is annotated with required knowledge fact. Recently, benchmarks have been set up that focus on open-domain knowledge [18], [19], which means that KBs are not supplied and answers can be derived from any external knowledge source.

• This work is supported by the National Natural Science Fund of China (62221002, 62171183) , the Hunan Provincial Natural Science Foundation of China (2022JJ20017), and partially sponsored by CAAI-Huawei MindSpore Open Fund. (Corresponding author: Shutao Li, Bin Sun.)
• Z. Ma, S. Li, B. Sun, Z. Long are with College of Electrical and Information Engineering, and with the Key Laboratory of Visual Perception and Artificial Intelligence of Hunan Province, Hunan University, Changsha, 410082, China. (maziyu@hnu.edu.cn; shutao_li@hnu.edu.cn; sunbin611@hnu.edu.cn; longzx@hnu.edu.cn)
• J. Cai is with the Data Science & AI Department, Faculty of IT, Monash University, Clayton, VIC 3800, Australia. (Jianfei.Cai@monash.edu)
• F. Ma is with Defense Innovation Institute, Chinese Academy of Military Science, Beijing, 100029, China. (mafuyan@hnu.edu.cn)

Early studies [20], [21], [22], [23], [24], [25], [26], namely retrieval-based paradigm, retrieve knowledge from multiple KBs (e.g., Wikipedia and ConceptNet [27]) and other knowledge sources for knowledge-based VQA. Afterwards, the retrieved knowledge passages, image, and question are sent to a multi-modal reasoning model for answer prediction. However, these retrieval-based methods have the following limitations: (i) the required knowledge may fail to retrieve from multiple KBs and other knowledge sources; (ii) even though the required knowledge is successfully retrieved, a lot of irrelevant information is ineluctably introduced meanwhile, which impairs the performance of their models.

Except for those researches using KBs, another paradigm uses a pre-trained large language model, e.g., GPT-3 [1], as an implicit knowledge engine to jointly acquire and reason the relevant knowledge to predict the answer directly. Given an image-question pair, these methods first translate the image into task-specific textual information (e.g., captions [28], [29] and answer heuristics [30]), facilitating the understanding of GPT-3 [31]. Then the question, task-specific textual information, and in-context examples are formatted as a task-specific textual prompt to induce GPT-3 to answer the question directly. Despite the promising performance achieved by these methods, such modality transformation methods may introduce irrelevant and harmful information to the LLM, which cause GPT-3 to misinterpret images and ignore visual details crucial for accurate answer. For example, the generic caption *'an airplane is parked in the middle of a jungle, surrounded by trees and foliage'* is useless to answer the question *'What type of trees are in this picture?'*. In this case, GPT-3 can only make a biased and aimless surmise

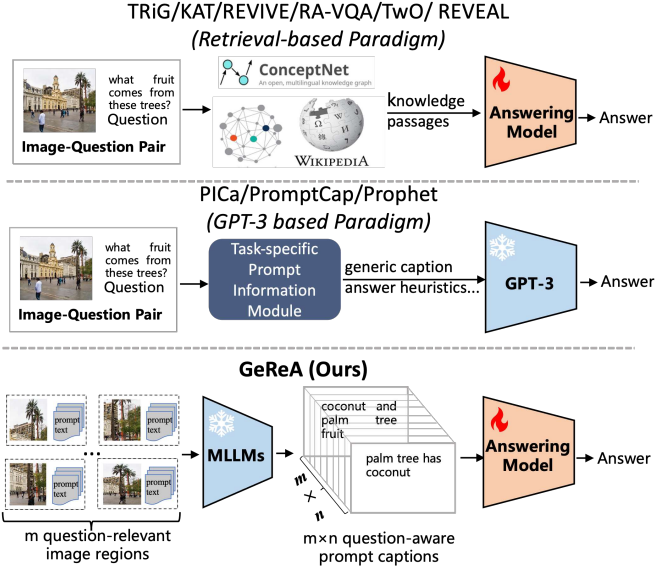

Fig. 1. Conceptual comparisons of three categories of knowledge-based VQA paradigms. The retrieval-based paradigm relies on open-domain knowledge bases and other knowledge sources to acquire knowledge. The GPT-3 based paradigm translates the image into task-specific textual information (e.g., captions and answer heuristics) to activate the capacity of GPT-3 to jointly acquire and reason the relevant knowledge for answering. Different from the two categories of paradigms, GeReA acquires the knowledge by prompting the MLLM with the question-relevant image regions and question-specific manual prompts to generate the question-aware prompt captions.

to predict the answer.

Multimodal large language models like InstructBLIP [2] have superior visual understanding capability compared to LLMs, by constructing a lightweight Querying Transformer between visual encoders (e.g., ViT-L/14 [32], ViT-g/14 [33]) and LLMs (e.g. Vicuna [34], FlanT5 [35]). Moreover, in the knowledge-based VQA task, InstructBLIP achieves 57.62%[1] zero shot performance on OK-VQA dataset [18], outperforming many knowledge-based VQA methods [21], [22], [23], [36], [37]. We argue that MLLM might be more suitable as implicit knowledge engine than LLM. However, how to activate the capacity of MLLM as an implicit knowledge engine to acquire more relevant and richer knowledge for knowledge-based VQA remains unknown.

To solve the above problem, we turn to how human use knowledge to answer visual question. As a human to solve this task, one needs to first understand the question, and locate the question relevant regions in the image. Then, the human needs to associate the question relevant image regions with the question prompts to activate relevant knowledge. Finally, the human answers the question with the relevant knowledge. Inspired by this, the question relevant image regions and the question relevant prompts are both fed to the MLLM (e.g., InstructBLIP) to obtain knowledge-relevant descriptions, namely question-aware prompt captions. The technology details of the question relevant image regions and the question relevant prompts are presented in Section 3.1. By observing the output captions, we find that the answer to the question is contained within these captions in most cases. Furthermore, we conduct experiments on OK-VQA dataset and surprisingly find the answer hit rate of this caption method reaches up to 86% in both the training and test sets. The results imply that prompting the MLLM as the implicit knowledge engine with question-relevant vision and language information is feasible.

Given a large amount of question-aware prompt captions, the next challenge is how to correctly predict the answer from these captions, which may include some irrelevant and erroneous information. Due to the length limit of LLM (i.e., 2048 in GPT-3), it cannot leverage the strong reasoning capability of GPT-3 to predict the answer. Therefore, we present to learn a strong joint knowledge-image-question representation from these question-aware prompt captions for final answer prediction.

In this paper, we propose GeReA, a generate-reason framework for knowledge-based VQA. Different from the retrieval based and GPT-3 based methods, this framework acquires knowledge by generating question-aware prompt captions, which are obtained by prompting a MLLM (e.g., InstructBLIP) with question relevant vision and language information. It has two stages, i.e., question-aware prompt caption generation and question-aware prompt caption reasoning. In the question-aware prompt caption generation stage, the top-$m$ most question relevant image regions are selected by the image-text matching. The text prompts are manually constructed using question-relevant information, namely question-specific manual prompts. The question-relevant image regions and question-specific manual prompts are both encoded in the frozen MLLM to generate the question-aware prompt captions. In the question-aware prompt caption reasoning stage, the question-aware prompt captions, image-question pair, and similar samples are input into the multi-modal reasoning model to learn a strong joint knowledge-image-question representation for answer prediction. Interestingly, the MLLM used in GeReA is replaceable and the performance of the proposed framework directly depends on the zero shot performance of the MLLM in knowledge-based VQA task. A simple schematic of GeReA is shown in Fig. 1(c).

GeReA obtains a consistent performance gain (4.48% and 4.52% on OK-VQA) over the corresponding MLLM (i.e., InstructBLIP, LLaVA-1.5 [38]) that the zero shot performance is 57.62% and 59.10%[2] on OK-VQA respectively. Combining the captions from different MLLMs (InstructBLIP and LLaVA-1.5), GeReA obtains state-of-the-art result on OK-VQA dataset (66.46%). It outperforms all previous state-of-the-art methods, even the PaLM-E [39] model with 562 billion parameters.

---

[1]. To evaluate the zero shot performance on OK-VQA dataset for InstructBLIP, we use the text prompt 'question: {question} the answer: ' and beam search for decoding, where sampling temperature is set to 0. The text prompt is used for VQA datesets in original paper. Note that the checkpoints of the used InstructBLIP is not fine-tuned on OK-VQA dataset due to it is not open source. The checkpoints are available at 'https://storage.googleapis.com/sfr-vision-language-research/LAVIS/models/InstructBLIP/instruct_blip_vicuna7b_trimmed.pth'

[2]. To evaluate the zero shot performance on OK-VQA dataset for LLaVA-1.5, we use the text prompt '{question}\n answer the question using a single word or phrase.' and beam search for decoding, where sampling temperature is set to 0. The text prompt is used for VQA datasets in original paper. The checkpoints are available at 'https://huggingface.co/liuhaotian/llava-v1.5-7b'

Furthermore, we explore the generalization ability of our GeReA on A-OKVQA [19], indicating that GeReA achieves a 4.61% improvement over its corresponding MLLM and outperforms all existing state-of-the-art methods.

To summarize, this work has the following contributions:

1) We propose a new paradigm for the knowledge-based VQA task. Different from the previous retrieval based and GPT-3 based paradigm, the proposed paradigm first prompts the MLLM with question relevant vision and language information to acquire relevant knowledge and then learns a strong joint knowledge-image-question representation for answer prediction. This is the first study using the MLLM as the implicit knowledge engine for knowledge-based VQA without extra knowledge bases.
2) Without bells and whistles, GeReA outperforms all existing state-of-the-art methods on OK-VQA and A-OKVQA datasets, with testing accuracies of 66.5% and 63.3%, respectively.

The rest of our work is structured as follows: The related work is presented in Section 2. Our GeReA is described in Section 3. The experimental settings and the results are shown in Section 4. Section 5 analyzes and discusses the experimental results. Finally, we make the conclusive remarks in Section 6.

## 2 RELATED WORK

### 2.1 Retrieval-based Methods

The key challenge of knowledge-based VQA task is how to accurately acquire the knowledge and integrate it with the image-question pair to predict the correct answer. Early researches [15], [18], [40] first transform the image-question pair into structured queries which aligns with the pre-defined knowledge bases (KBs), and then retrieve the required knowledge fact from these KBs to get the final answer. Due to the insufficiency of pre-defined knowledge to represent general knowledge, subsequent work pays attention to acquire knowledge from open-domain knowledge bases, such as Wikipedia [41], Google Images [23] and ConceptNet [27].

TRiG [22] proposes to align the visual information into the language space at three-level (i.e., caption [6], dense label [42] and OCR [43]), and then retrieves the relevant knowledge passages from Wikipedia [41] via dense passage retrieval (DPR). Considering separate training for DPR and answer generation might limit the overall performance, RA-VQA [37] presents a strong joint end-to-end training framework which includes a knowledge retriever and an answer generator for acquiring more relevant knowledge from Google Search [44]. Afterwards, to handle the two major deficiencies in RA-VQA's retriever: (i) the incomplete and inaccurate image representations (ii) insensitivity to finer-grained relevance, RA-VQA-v2 [45] proposes to incorporate complementary image representations from a vision model aligned with a text-based retriever and employ multi-dimensional embeddings for finer-grained relevance between queries and documents, significantly improving the knowledge retrieval performance of the original RA-VQA.

ReVeaL [26] presents an end-to-end pre-training paradigm for knowledge-based VQA. Different from RA-VQA, ReVeaL gets relevant knowledge by constructing a large-scale memory, which obtained by encoding various knowledge sources, including knowledge graph [46], Wikipedia passage [41], and web images with alt-text captions [47].

Motivated by the strong knowledge retrieval ability of LLM (e.g., GPT-3 [1]), KAT [24] and REVIVE [25] introduce two types of knowledge sources, i.e., Wikipedia [41] and GPT-3 [1]. These works derive potential answers along with supporting evidence from GPT-3 and retrieve knowledge passages from Wikipedia via DPR. Afterwards, TwO [48] introduces four specific types of knowledge sources to acquire knowledge, including textual knowledge from Wikipedia [41] and GPT-3 [1], multimodal knowledge from VQAv2 dataset [49] and OFA [50]. Even though the introduction of large language models to extract potential answers with evidence, they still highly rely on open-domain knowledge bases to acquire knowledge.

### 2.2 Large Language Model Based Methods

Except for those researches using KBs, another paradigm directly prompts pre-trained large language models, e.g., GPT-3 [1] to jointly acquire and reason the relevant knowledge for answering. PICa [28] is the pioneering study and explores the frozen GPT-3 model with textual prompt as its input to predict the answer directly. Specifically, PICa first converts the image into task-specific textual information (e.g., generic caption) via an VinVL-base [51] model. Then the question, task-specific textual information, and in-context examples are formatted as a task-specific textual prompt to induce GPT-3 to answer the question directly.

Thanks to the successful attempts of PICa and the powerful knowledge retrieval and reasoning abilities of GPT-3 [1], subsequent works are centered on endow GPT-3 with some task-specific information to improve its ability for knowledge-based VQA. In order to acquire more task-specific information, Prophet [30] proposes a novel and effective framework that converts images into answer-aware examples and answer candidates, significantly activating the capacity of GPT-3 for knowledge-based VQA. PromptCap [29] designs a task-aware captioning model to better connect the GPT-3 and images.

In addition, there are some impressive work based on large language model for zero shot knowledge-based VQA. PnP-VQA [52] proposes a modular framework without training that converts the image into question-relevant caption to activate the capacity of UnifiedQAv2 [53] for knowledge-based VQA. Given that PnP-VQA does not harness the advantages of in-context learning, Img2LLM [54] translates the image into multiple synthetic question-answer pairs which include question-relevant captions via a BLIP [55] model and a question-generation model to further activate the capacity of LLM under the zero shot setting. The two works both use the BLIP model to generate the captions via the question-relevant portion of the image and an image-text matching model to filter the noisy captions. Then they use the captions as textual prompt to activate the capability of LLM as an implicit knowledge engine for knowledge-based VQA. The BLIP model is a modality translator and the knowledge acquisition and integration highly rely on the LLM.

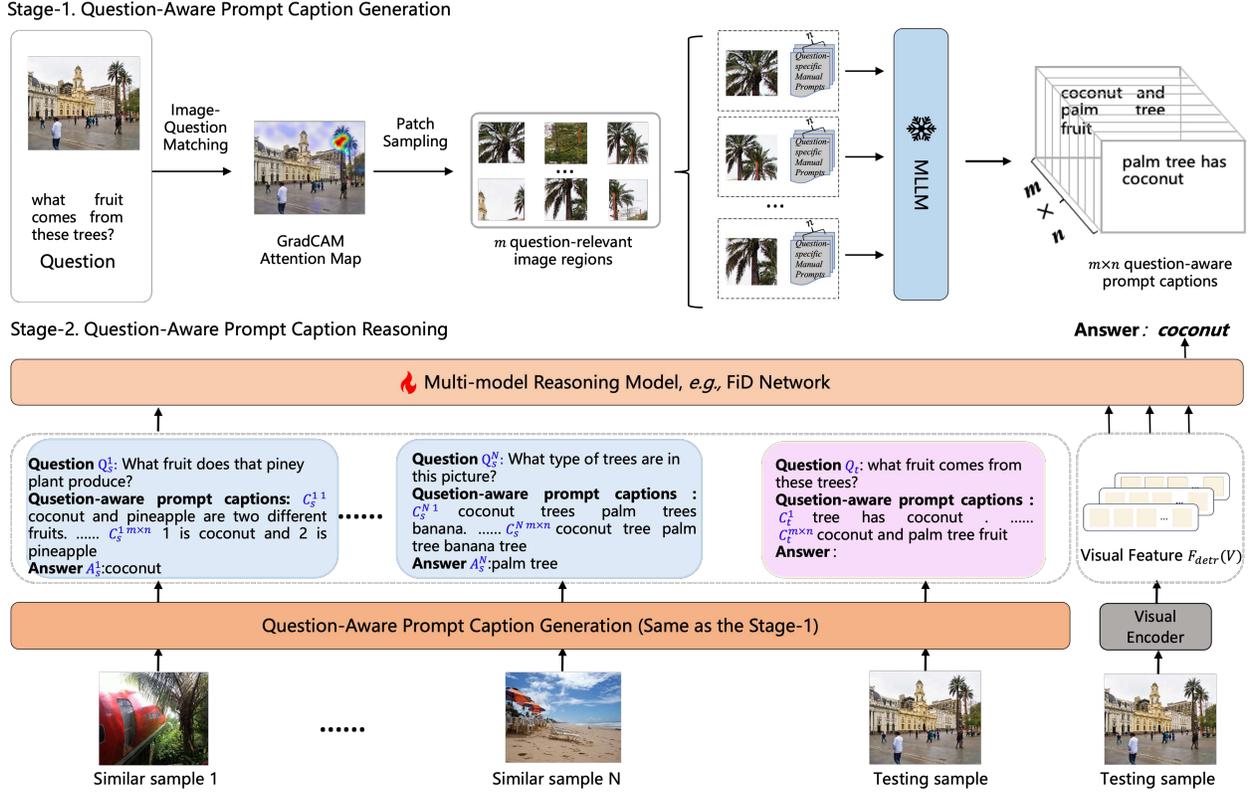

Fig. 2. The architecture of our GeReA. In the question-aware prompt caption generation stage, the question-relevant image regions and question-specific manual prompts are input into the frozen MLLM to generate $m \times n$ question-aware prompt captions. In the question-aware prompt caption reasoning stage, the question-aware prompt captions, visual feature, and similar samples are integrated into the multi-modal reasoning model to predict the final answer.

Different from the two works, GeReA explores to activate the capacity of the MLLM as an implicit knowledge engine to get more relevant and richer knowledge for knowledge-based VQA. The acquisition of knowledge is achieved by prompting the MLLM (e.g., InstructBLIP) with two types of task-specific information, i.e., the question relevant image regions and question-specific manual prompts to generate question-aware prompt captions. The integration of knowledge is achieved by learning a knowledge-image-text representation from the question-aware prompt captions, image-text pair, and similar samples via a multi-modal reasoning model. Moreover, the answer hit rate of PnP-VQA [52] and Img2LLM [54] is 58.89%, while the answer hit rate of our question-aware prompt captions is 86.36% on OK-VQA dataset under the same MLLM (i.e., InstructBLIP).

### 2.3 Pre-Training Methods

Multimodal Large Language Model (MLLM) leverages the powerful LLM to perform multimodal tasks, demonstrating impressive performance such as visual knowledge retrieval and visual question answer. Some representative works include Flamingo [56], BLIP-2 [57], InstructBLIP [2], LLaVA [58], MiniGPT-4 [59], mPLUG-Owl [60], mPLUG-Owl2 [61], LLaMA-Adapter-v2 [62], Otter [63], Multimodal-GPT [64], PandaGPT [65], PaLI [66], PaLI-X [67], and PaLM-E [39].

These works align vision and language embeddings by pre-training a neural network on a large-scale image-text dataset. However, huge computation and data requirements are severely unaffordable for most researchers. For example, PaLI-17B requires 1,024 TPUv4 for 7 days and requires 1566M corpus for pre-training [66].

## 3 METHODOLOGY

In this section, we introduce the details of our proposed generate-reason framework for knowledge-based VQA. As shown in Figure 2, GeReA consists of two stages: (i) question-aware prompt caption generation, (ii) question-aware prompt caption reasoning. In the first stage, the task-specific vision and language information are fed to the frozen MLLM model to generate question-aware prompt captions. In the second stage, a caption-augmented VQA model is trained to learn a joint multi-modal representation to predict the final answer.

### 3.1 Question-Aware Prompt Caption Generation

Different from the previous retrieval based and GPT-3 based studies, GeReA acquires the knowledge via the frozen MLLM to generate the question-aware prompt captions. Two types of task-specific information, i.e., the question relevant image regions and question-specific manual prompts are used to activate the capacity of MLLM for knowledge retrieval. For the question relevant image regions, the off-the-shelf Image grounded Text Encoder (ITE) in BLIP-2 is used to calculate the similarity score $sim(V,Q)$ between image $V$ and question $Q$. Then the GradCAM generates a localisation map to determine top-$m$ most question relevant image regions [52]. For the question-specific manual prompts, six text prompt templates are constructed manually according to the question-relevant information.

TABLE 1
The List of Question-specific Manual Prompts for InstructBLIP

| Num | Prompt Template |
|---|---|
| 1) | write down the facts that you know about this picture |
| 2) | explain this picture in as much detail as possible based on the information provided below: {noun phrases and verb phrases} |
| 3) | {question} |
| 4) | question: {question} the answer: |
| 5) | question: {question} according to the question and image, we know that |
| 6) | explain this picture according to the question {question} |

TABLE 2
The List of Question-specific Manual Prompts for LLaVA-1.5

| Num | Prompt Template |
|---|---|
| 1) | write down the facts that you know about this picture |
| 2) | {noun phrases and verb phrases} \n explain this picture in as much detail as possible based on the provided information. |
| 3) | {question}\n answer the question using a single word or phrase. |
| 4) | question: {question} the answer: \n answer the question using a single word or phrase. |
| 5) | question: {question}\n according to the question and image, we know that |
| 6) | {question} \n explain this picture according to the question |

### 3.1.1 Question-Relevant Image Regions

We use the ITE to calculate a cross-attention matrix $W$ between image patch features and the corresponding question features. The cross-attention matrix $W$ reflects the patch importance when using the ITE to compute the similarity between the whole image and the question, but the cross-attention matrix contains much redundancy and some cross-attention scores are useless. Therefore, in accordance with GradCAM [68], we calculate the derivative of the similarity score $sim(V,Q)$ with respect to the cross-attention score, denoted as $sim(V,Q)/W$. Then an element-wise multiplication of its gradient matrix with the cross-attention matrix is performed to filter these useless attention scores [52].

Specifically, let $V_f^i \in \mathbb{R}^{M \times D_v^i}$ denote the image patch features obtained by ITE, $Q_f^i \in \mathbb{R}^{L \times D_q^i}$ denote the corresponding question features, where $M$ represents the total number of image patches, $i$ represents the layer number of ITE, $D_v^i$ represents the image patch feature dimension in the $i$-th layer of ITE, $L$ represents the token number of the corresponding question, $D_v^q$ represents the question feature dimension in the $i$-th layer of ITE. For the image patch features $V_f^i$ and the corresponding question features $Q_f^i$, their cross-attention scores $W^i$ between each question token and each image patch are computed with a cross-attention head in $i$-th layer as

$$W^i = \text{softmax}\left(\frac{Q_f^i W_{query}^i W_{key}^{i\top} V_f^{i\top}}{\sqrt{D_q^i}}\right) \quad (1)$$

where $W_{query}^i \in \mathbb{R}^{D_q^i \times D_q^i}$ and $W_{key}^i \in \mathbb{R}^{D_v^i \times D_q^i}$ represent the query head and the key head respectively in the cross-attention head of the $i$-th ITE layer. After obtaining the cross-attention matrix $W^i \in \mathbb{R}^{L \times M}$, we compute the derivative of $sim(V,Q)/W_i$, and the element-wise multiplication between the cross-attention matrix and its gradient matrix is performed. Finally, we sum over $L$ textual tokens of the corresponding question and average over $H$ attention heads to determine the $j$-th image patch relevance $r_j^i$:

$$r_j^i = \frac{1}{H}\sum_{l=1}^{L}\sum_{h=1}^{H}\min\left(0, \frac{\partial \, sim(V,Q)}{\partial W_{lj}^{ih}}\right) W_{lj}^{ih} \quad (2)$$

where $i$ represents the layer number of ITE, $h$ represents the number of attention heads.

After obtaining the image patch relevance $r$, we sample a subset of image patches $K$ with probability proportional to $r$. For each image, we repeat this $m$ times to determine the top-$m$ most question relevant image regions.

### 3.1.2 Question-Specific Manual Prompts

In order to fully explore the ability of MLLM, we conduct the prompts with four perspectives for each question relevant image region. The first class is the unconstrained prompt giving the MLLM a task to describe the image region without any other information including question, *e.g., write down the facts that you know about this picture*. The second class is the noun-verb phrases prompt which extracts noun phrases (including named entities) and verb phrases from the question, making the MLLM associate with the task-relevant context, *e.g., explain this picture in as much detail as possible based on the information provided below: {noun phrases and verb phrases}*. The third class is the question inquiry prompt which asks the question to the MLLM directly and inherits the zero shot capability of the MLLM on the knowledge-based VQA task, *e.g., question: {question} the answer:* . The final class is the question-relevant description prompt that gives the MLLM a task to describe the image region about question information, *e.g., question: {question} according to the question and image, we know that* . Since multiple instruction templates are used for converting the used VQA and image captioning datasets into instruction tuning data both in InstructBLIP and LLaVA-1.5, we conduct two manual templates for the question inquiry prompt and the question-relevant description prompt, respectively.

Finally, $n = 6$ question-specific prompts are conducted by humans, where $n$ denotes the number of prompts. Due to the different response formatting prompts during the training between InstructBLIP and LLaVA-1.5, we make a slight change to the question-specific prompts. Table 1 shows the question-specific prompts to InstructBLIP, while Table 2 shows the question-specific prompts to LLaVA-1.5.

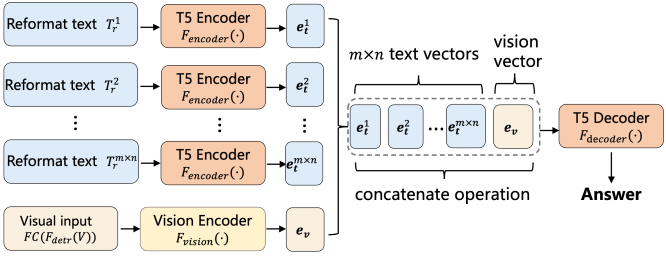

Fig. 3. The architecture of the multi-modal reasoning model.

TABLE 3
Model configuration of the MLLM used in GeReA

| MLLMs | Vision Encoder | LLMs | # params |
|---|---|---|---|
| InstructBLIP | ViT-g/14 | Vicuna | 7B |
| LLaVA-1.5 | ViT-L/14@336px | Vicuna | 7B |

For each image-text pair, the top-$m$ most question relevant image regions and $n$ question-specific manual prompts are encode to the frozen MLLM to generate $m \times n$ question-aware prompt captions.

### 3.2 Question-Aware Prompt Caption Reasoning

Given a large amount of question-aware prompt captions, the next challenge is how to correctly predict the answer from these captions, which may include some irrelevant and erroneous information.

To tackle this, we propose to train a multi-modal reasoning model to learn the joint knowledge-image-text representation for answer prediction. Specially, the FiD [69] network structure is adopted as the multi-modal reasoning model to reason the generated question-aware prompt captions. The FiD network is a variant of the T5 model [70]. It consists of multiple T5 encoders and a T5 decoder. The FiD network is widely used in knowledge-based VQA task. Many methods such as TRiG [22], KAT [24], REVIVE [25], and TwO [48] use the same network for answer prediction.

To better learn the joint representation, the relevant information, i.e., image-text pair and similar samples is also input into the FiD network to assist in reasoning. The image-text pair involves the question and the visual feature which extracted by the DETR encoder [71]. The similar samples are selected from the training set of the knowledge-based VQA datasets. In this paper, we follow the prophet method [30] to extract the similar samples. The architecture of the multi-modal reasoning model is shown in Fig. 3

Due to the length limitation of the FiD network, it is impossible to encode all the captions of the testing and similar samples with one encoder. Therefore, we use $m \times n$ encoders to encode these captions. Specifically, let $Q_t$ denote question of the testing sample, $C_t = \{C_t^j | j = 1, \ldots m \times n\}$ denote the question-aware prompt captions of the testing sample, $C_s^n = \{C_s^{nj} | j = 1, \ldots m \times n\}$ denote the $n$-th question-aware prompt captions of the similar samples, where $n = 1, \ldots N$. $Q_s = \{Q_s^n | n = 1, \ldots N\}$ denote the question of the similar samples, $A_s = \{A_s^n | n = 1, \ldots N\}$ denote the answer of the similar samples. For $j$-th encoder of FiD network, the input text is reformatted as follows:

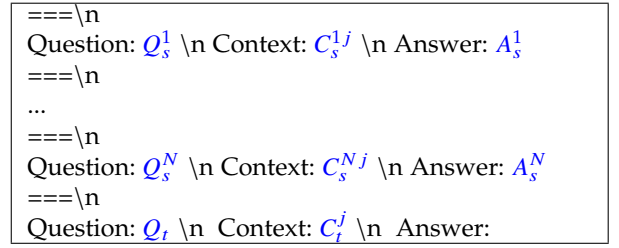

We denote the reformat text as $T_r^j$, where $j = 1, \ldots m \times n$. Then the reformat text is input into the FiD encoder $F_{encoder}$,

$$\mathbf{e}_t^j = F_{encoder}\left(T_r^j\right) \tag{3}$$

where $\mathbf{e}_t^j \in \mathbb{R}^D$ and $D$ means the embedding dimension.

Besides the text feature, the DETR encoder which denotes as $F_{detr}(\cdot)$ is used to extract the visual feature of the testing sample. Subsequent, the visual feature feeds to a transformer [72], [73] encoder $F_{vision}$,

$$\mathbf{e_v} = F_{vision}\left(FC(F_{detr}(V))\right) \tag{4}$$

where $\mathbf{e_v} \in \mathbb{R}^D$ and $D$ means the embedding dimension which is similar to the $F_{encoder}$ encoder. $V$ represents the whole image, $FC$ is the fully-connected layer which used to align the dimension between the DETR encoder and the transformer encoder.

After encoding of all the input, i.e., the question-aware prompt captions, image-text pair, and similar samples, a concatenate operation is performed for these vectors along the first dimension. Then we sent the concatenated vector into the decoder $F_{decoder}$ of the FiD network to predict the answer,

$$y = F_{decoder}\left(\text{concat}\left(\mathbf{e}_t^1, \cdots, \mathbf{e}_t^{m \times n}, \mathbf{e_v}\right)\right) \tag{5}$$

where $y$ represents the generated answer. We use the cross entropy loss function to train the model, which is shown as below:

$$\mathcal{L} = -\sum_{\ell=1}^{L_{gt}} \log p_\theta\left(y_\ell \mid y_{<\ell}\right) \tag{6}$$

in which $L_{gt}$ represents the length of the human response, $y_\ell$ represents the human response at the position $\ell$, $\theta$ represents the parameters of the multi-modal reasoning model.

In addition, model ensemble is used to integrate results trained on multiple models to further improve the performance. Specifically, three models are trained with different initialized seeds, and the final result is selected by majority voting for each sample.

## 4 EXPERIMENT

To evaluate the validity of the proposed GeReA, extensive experiments are performed on OK-VQA and A-OKVQA datasets. In addition, the results of the proposed GeReA are compared with the retrieval-based methods, GPT-3 based methods and pre-training methods.

TABLE 4
Results comparison with existing methods on OK-VQA, with the knowledge source each method uses. The table is divided into three parts. The first part shows retrieval-based methods utilizing external KBs and other external knowledge sources. The second part presents GPT-3 based methods which use the GPT-3 as the only knowledge source to joint acquire and reason the relevant knowledge. The final part lists pre-training methods that train a neural network on a large-scale image-text dataset. * represents the results not reported in the original paper, re-implemented by ourselves following the instructions provided in the original paper. Please see page 2.

| # | Method | Publication | Knowledge Source | Accuracy (%) |
|---|---|---|---|---|
| | | *Methods with External Knowledge Bases* | | |
| 1 | BAN + KG + AUG [74] | MM'2020 | Wikipedia + ConceptNet | 26.7 |
| 2 | ConceptBERT [36] | EMNLP'2020 | ConceptNet | 33.7 |
| 3 | KRISP [21] | CVPR'2021 | Wikipedia + ConceptNet | 38.4 |
| 4 | Vis-DPR [75] | EMNLP'2021 | Google Search | 39.2 |
| 5 | MAVEx [23] | AAAI'2022 | Wikipedia + ConceptNet + Google Images | 39.4 |
| 6 | TRiG [22] | CVPR'2022 | Wikipedia | 50.5 |
| 7 | KAT (Single) [24] | NAACL'2022 | GPT-3 (175B) + Wikidata | 53.1 |
| 8 | KAT (Ensemble) [24] | NAACL'2022 | GPT-3 (175B) + Wikidata | 54.4 |
| 9 | RA-VQA [37] | EMNLP'2022 | Google Search | 54.5 |
| 10 | REVIVE (Single) [25] | NeruIPS'2022 | GPT-3 (175B) + Wikidata | 56.6 |
| 11 | TwO (Single) [48] | ACL'2023 | Wikipedia + Frozen OFA (0.93B) + VQA-2 + GPT-3 (175B) | 57.6 |
| 12 | REVIVE (Ensemble) [25] | NeruIPS'2022 | GPT-3 (175B) + Wikidata | 58.0 |
| 13 | TwO (Ensemble) [48] | ACL'2023 | Wikipedia + Frozen OFA (0.93B) + VQA-2 + GPT-3 (175B) | 58.7 |
| 14 | ReVeaL [26] | CVPR'2023 | WIT + CC12M + Wikidata + VQA-2 | 59.1 |
| 15 | RA-VQA-v2 [45] | NeruIPS'2023 | Google Search | 62.1 |
| | | *Methods with GPT-3 API (>175 Billion Parameters)* | | |
| 16 | PICa-Base [28] | AAAI'2022 | GPT-3 (175B) | 43.3 |
| 17 | PICa-Full [28] | AAAI'2022 | GPT-3 (175B) | 48.0 |
| 18 | PromptCap [29] | ICCV'2023 | GPT-3 (175B) | 60.4 |
| 19 | Prophet [30] | CVPR'2023 | GPT-3 (175B) | 61.1 |
| | | *Pre-Training Methods* | | |
| 20 | PaLI-3B [66] | ICLR'2023 | PaLI (3B) | 52.4 |
| 21 | PaLI-15B [66] | ICLR'2023 | PaLI (15B) | 56.5 |
| 22 | InstructBLIP-7B (Generalist) [2] | NeruIPS'2023 | InstructBLIP-7B (Generalist) | 57.6* |
| 23 | Flamingo-80B [56] | NeruIPS'2022 | Flamingo (80B) | 57.8 |
| 24 | LLaVA-1.5-7B (Generalist) [38] | - | LLaVA-1.5-7B (Generalist) | 59.1* |
| 25 | PaLM-E-12B (Single-task Finetuned) [39] | ICML'2023 | PaLM-E (12B) | 60.1 |
| 26 | InstructBLIP-7B (Single-task Finetuned) [2] | NeruIPS'2023 | InstructBLIP-7B (Single-task Finetuned) | 62.1 |
| 27 | PaLM-E-66B (Single-task Finetuned) [39] | ICML'2023 | PaLM-E (66B) | 62.9 |
| 28 | PaLM-E-84B (Single-task Finetuned) [39] | ICML'2023 | PaLM-E (84B) | 63.3 |
| 29 | PaLI-17B [66] | ICLR'2023 | PaLI (17B) | 64.5 |
| 30 | PaLM-E-562B (Generalist) [39] | ICML'2023 | PaLM-E (562B) | 66.1 |
| 31 | PaLI-X-55B (Single-task Finetuned) [67] | - | PaLI-X (55B) | 66.1 |
| | | *Ours* | | |
| 32 | GeReA (Single) | - | InstructBLIP-7B (Generalist) | 62.1 |
| 33 | GeReA (Single) | - | LLaVA-1.5-7B (Generalist) | 63.6 |
| 34 | GeReA (Single) | - | InstructBLIP-7B (Generalist) + LLaVA-1.5-7B (Generalist) | 65.4 |
| 35 | GeReA (Ensemble) | - | InstructBLIP-7B (Generalist) + LLaVA-1.5-7B (Generalist) | **66.5** |

### 4.1 Datasets

OK-VQA is a prevalent open-domain knowledge-based VQA dataset [18] which contains 9K training image-question pairs and 5K testing image-question pairs. Every question requires the external knowledge to answer. Each data sample includes ten open-ended human responses as annotations. The 1.1 version of the dataset is used to conduct the below experiments.

A-OKVQA [19] is an augmented successor of OK-VQA and is divided into three parts: 17K image-question pairs for training, 1K image-question pairs for validation, and 7K image-question pairs for testing. It contains two task, i.e., direct answer (DA) task and Multiple Choice (MC) task. For the DA task, Each question includes ten open-ended human responses as annotations. For the MC task, it requires model to select the correct answer from four options.

### 4.2 Evaluation

The accuracy calculated by the VQA score is the evaluation metric which assigns score to the predicted answer based on its exact count of occurrences within the set of human responses $S$:

$$\text{Accuracy}(y, S) = \min\left(\frac{\#S(y)}{3}, 1\right) \qquad (7)$$

where $\#S(y)$ represents the occurrence of $y$ in human responses $S$. This scoring method ensures a model receives partial credit, even if it generates an answer less popular among human responses.

TABLE 5
Results comparison with existing methods on A-OKVQA. Both tasks are measured by accuracy(%). * represents the results not reported in the original paper, re-implemented by ourselves following the instructions provided in the original paper. Please see page 9.

| Method | Direct Answer | | Multiple Choice | |
| --- | --- | --- | --- | --- |
| | val | test | val | test |
| ClipCap *(ECCV 2022)* [19] | 18.1 | 15.8 | 44.0 | 43.8 |
| Pythia *(CVPR 2018)* [76] | 25.2 | 21.9 | 49.0 | 40.1 |
| ViLBERT *(NeruIPS 2019)* [8] | 30.6 | 25.9 | 49.1 | 41.5 |
| LXMERT *(EMNLP 2019)* [77] | 30.7 | 25.9 | 51.4 | 41.6 |
| KRISP *(CVPR 2021)* [21] | 33.7 | 27.1 | 51.9 | 42.2 |
| GPV-2 *(ECCV 2022)* [78] | 48.6 | 40.7 | 60.3 | 53.7 |
| Prophet *(CVPR 2023)* [30] | 58.2 | 55.7 | 76.4 | 73.6 |
| LLaVA-1.5-7B (Generalist) *(arXiv 2023)* [38] | 63.7* | 58.6* | 77.1* | 74.5* |
| InstructBLIP-7B(Generalist) *(NeruIPS 2023)* [2] | 62.4* | 58.7* | 73.0* | 71.1* |
| PromptCap *(ICCV 2023)* [29] | 56.3 | 59.6 | 73.2 | 73.1 |
| InstructBLIP-7B(Single-task Finetuned) *(NeruIPS 2023)* [2] | 64.1 | 62.1 | 75.7 | 73.4 |
| **GeReA (Single)** (InstructBLIP+LLaVA-1.5-7B) | **68.0** | **63.3** | **79.7** | **75.9** |

## 4.3 Implementation Details

In the question-aware prompt caption generation stage, two different MLLMs i.e., InstructBLIP and LLaVA-1.5, are used to generate the question-aware prompt captions, whose configurations are shown in Table 3. The vision encoder ViT-g/14 and the LLM Vicuna (7B) are used in InstructBLIP. The 6-*th* cross-attention layer of ITE is selected for computing the cross-attention matrix $W$. We set the subset of image patch $K = 20$ and repeat $m = 20$ times for each image to identify the 20 most relevant image regions. The length of question-aware prompt caption ranges from 10 to 25. We use beam search for decoding, where top = 0.9, temperature = 1.0, num_beams = 5. When LLaVA-1.5 is used instead of InstructBLIP, the subset of image patch $K = 30$, and the length of question-aware prompt caption ranges from 1 to 32.

In the question-aware prompt caption reasoning stage, 2 × NVIDIA A100 40GB are used to train the multimodal reasoning model for 20k steps with 2 days. AdamW [79] is the chosen optimizer and the learning rate is set to 5e5. The pre-trained T5 model [70], i.e., T5-large is used to initialize the multimodal reasoning model, which is same as the KAT [24]. The batch size is 1, the warm-up steps are 1K, and the multimodal reasoning model is evaluated every 10000 steps during training. We evaluate the prediction results post-normalization, which primarily involves removing punctuation, duplicated white space, and converting to lower case [80].

## 4.4 Main Results

Due to the different length of question-aware prompt caption from different MLLMs and the length limit of T5-large model, we set the number of similar sample $N = 10$ as our setting for InstructBLIP and set the number of similar sample $N = 5$ as our setting for LLaVA-1.5 and InstructBLIP+LLaVA-1.5[3] for the comparisons below.

---

3. We select the first 80 captions in LLaVA-1.5 and the first 40 captions in InstructBLIP for each sample as the knowledge source

Table 4 presents a comparison of our GeReA with existing state-of-the-art methods on the challenging OK-VQA dataset. The table is divided into three parts. The first part shows retrieval-based methods utilizing external KBs and other external knowledge sources. The second part presents GPT-3 based methods which use the GPT-3 as the only knowledge source to joint acquire and reason the relevant knowledge. The final part lists pre-training methods that train a neural network on a large-scale image-text dataset.

Our GeReA does not belong to either of the three parts which acquires the knowledge via the MLLM to generate question-aware prompt captions. GeReA uses three types of knowledge sources, i.e., InstructBLIP, LLaVA-1.5, and InstructBLIP + LLaVA-1.5 and achieves 62.1%, 63.6%, and 65.4% accuracies on OK-VQA respectively. Especially our ensemble model which uses the blended knowledge source, i.e., InstructBLIP + LLaVA-1.5 achieves 66.5% accuracy on OK-VQA, outperforming all the compared methods, including the heavily-engineered PaLM-E model with 562 billion parameters. We also find that GeReA gives a consistent performance gain (4.48% and 4.52% on OK-VQA) over the corresponding MLLM (i.e., InstructBLIP, LLaVA-1.5) whose zero shot performance is 57.62% and 59.10% on OK-VQA respectively. It indicates that the MLLM used in our GeReA is replaceable and our GeReA can fully activate the capacity of MLLM as the implicit knowledge engine to acquire relevant knowledge for knowledge-based VQA. This reflecting the scalability and flexibility of our GeReA. Moreover, combining the captions from different MLLMs can enhance the performance of our GeReA on OK-VQA dataset. This is because the quality of captions becomes higher through this way. The detail is discussed in the next section.

Compared with the previous state-of-the-art retrieval-based methods, ReVeaL [26] and RA-VQA-v2 [45], GeReA is much simpler and gains at least 4.4% improvement on OK-VQA dataset. It is 8.5% accuracy higher than REVIVE [25] when both methods use the same multi-modal reasoning model for the final answer prediction. This indicates that harnessing the knowledge retrieval capabilities of the MLLM has more potential than retrieving knowledge from databases. Compared with the previous state-of-the-art GPT-3 based methods, PromptCap [29] and Prophet [30], our GeReA also surpasses all the methods on OK-VQA dataset with at least 5.4% improvement. Moreover, GeReA has at most 20 billion parameters, while the GPT-3 based methods exceed 175 billion parameters. This can be seen as an evidence that MLLM is more suitable as the implicit knowledge engine than LLM for knowledge-based VQA. Compared with the pre-training methods, GeReA is more resource-efficient and outperforms all the existing pre-training methods. When the parameter sizes are similar to PaLI-17B, our GeReA improves over it by 2% accuracy on OK-VQA dataset. Besides, GeReA has only 3.6% parameters compared to PaLM-562B, achieving higher performance on OK-VQA. It indicates that GeReA observably decreases the cost and more effective on the knowledge-based VQA task.

Table 5 shows the contrastive results on the other challenging dataset, i.e., A-OKVQA. We compare our GeReA to the baselines listed in [19], the MLLMs (i.e., InstructBLIP[4] [2], LLaVA-1.5[5] [38]) and existing state-of-the-art methods [2], [29], [30]. GeReA uses the blended knowledge source, i.e., InstructBLIP + LLaVA-1.5[6] and achieves state-of-the-art performance in both direct answer and multiple choice tasks. Specifically, GeReA achieves a significant improvement (4.6% on testing set in direct answer task) over its corresponding knowledge source (InstructBLIP + LLaVA-1.5[7]) and outperforms all the existing state-of-the-art methods in both tasks, showing the generalization and effectiveness of our GeReA.

TABLE 6
Per-category accuracies of the MLLM and GeReA. This performance improvements of our GeReA are observed on almost all categories. ZSP denotes zero shot performance, Acc. represents accuracy. For each category, VT represents Vehicles and Transportation; BCP represents Brands, Companies and Products; OMC represents Objects, Material and Clothing; SR represents Sports and Recreation; CF represents Cooking and Food; GHLC represents Geography, History, Language and Culture; PEL represents People and Everyday Life; Plants and Animals (PA); ST represents Science and Technology; WC represents Weather and Climate; and Other.

| Category | InstructBLIP | | | | LLaVA-1.5 | | | | InstructBLIP +LLaVA-1.5 | | | |
|---|---|---|---|---|---|---|---|---|---|---|---|---|
| | ZSP | AHR | ANR | Acc. | ZSP | AHR | ANR | Acc. | ZSP | AHR | ANR | Acc. |
| VT | 55.51 | 85.12 | 65.78 | 56.74 | 55.76 | 79.00 | 55.35 | 58.30 | 55.76 | 87.45 | 58.98 | 61.79 |
| BCP | 56.19 | 82.56 | 68.79 | 58.91 | 61.82 | 80.81 | 52.47 | 62.98 | 61.82 | 88.95 | 58.10 | 66.47 |
| OMC | 57.32 | 90.65 | 58.71 | 60.59 | 55.61 | 82.48 | 50.13 | 61.99 | 55.61 | 91.82 | 53.13 | 62.77 |
| SR | 55.85 | 88.36 | 60.99 | 67.31 | 61.55 | 82.89 | 49.56 | 67.72 | 61.55 | 90.65 | 53.63 | 70.08 |
| CF | 59.54 | 86.34 | 64.30 | 64.31 | 60.04 | 83.48 | 52.50 | 66.75 | 60.04 | 89.57 | 56.71 | 68.49 |
| GHLC | 58.63 | 81.56 | 67.39 | 63.36 | 64.54 | 74.47 | 55.37 | 65.01 | 64.54 | 85.11 | 59.70 | 66.43 |
| PEL | 55.30 | 81.78 | 64.88 | 60.28 | 58.41 | 77.10 | 57.52 | 61.92 | 58.41 | 85.28 | 60.33 | 66.67 |
| PA | 60.19 | 88.02 | 64.04 | 61.94 | 58.53 | 77.56 | 56.12 | 62.29 | 58.53 | 89.53 | 59.10 | 65.39 |
| ST | 51.98 | 84.52 | 67.63 | 53.97 | 61.11 | 77.38 | 53.44 | 57.54 | 61.11 | 88.10 | 58.09 | 61.11 |
| WC | 56.33 | 89.15 | 64.53 | 69.51 | 66.93 | 79.84 | 52.53 | 70.28 | 66.93 | 89.92 | 57.07 | 74.42 |
| Other | 58.97 | 86.11 | 62.65 | 62.20 | 59.83 | 81.58 | 53.63 | 65.21 | 59.83 | 89.50 | 56.90 | 66.51 |
| Overall | 57.62 | 86.39 | 63.80 | 62.10 | 59.10 | 80.28 | 53.70 | 63.62 | 59.10 | 89.00 | 57.33 | 65.41 |

## 5 ANALYSIS

To investigate the reasons for the improvements, we first make in-depth statistical analyses of GeReA's performance on the testing set of OK-VQA dataset, and then we conduct ablation experiments of each component. Moreover, six different types of cases with results of different methods are illustrated to demonstrate the effectiveness of our GeReA.

### 5.1 Capability of the MLLM

GeReA gives a performance gain (4.48%, 4.52% and 7.36% on OK-VQA) over its corresponding knowledge source (i.e., InstructBLIP, LLaVA-1.5, and InstructBLIP+LLaVA-1.5[8]). To better understand this improvement, we introduce two metrics, i.e., answer hit rate and answer noise rate to evaluate the quality of the generated question-aware prompt captions. Different from the Img2LLM [54], the answer hit rate (AHR) and answer noise rate (ANR) in our paper are defined as below:

$$AHR = \frac{\sum_{j=1}^{N_t} \min\left(\#TC^j(A), 1\right)}{N_t} \quad (8)$$

$$ANR = 1 - \frac{\sum_{j=1}^{N_t} \sum_{i=1}^{C_n} \min\left(\#TC_i^j(A), 1\right)}{N_t * C_n} \quad (9)$$

where $N_t$ represents the number of the testing samples, $C_n$ is the caption number of each testing sample, $A$ represents the ground-truth answer of the corresponding sample, $\#TC^j(A)$ is the occurrence of $A$ in $j$-th question-aware prompt caption of the testing sample. Moreover, we conduct two types of statistic analyses, i.e., per-category accuracies, and the prediction behaviors of GeReA.

#### 5.1.1 Per-category accuracies

Table 6 shows per-category performance of MLLM and GeReA with three different knowledge sources. The observations derived from the results are shown below: (i) When LLaVA-1.5 is used as the knowledge source instead of InstructBLIP, GeReA gets higher performance on all categories. The reason may be that the answer noise rate of LLaVA-1.5 is distinctly lower than that of InstructBLIP, which results in more captions for our GeReA that contain the answer, although the answer hit rate of InstructBLIP is higher than that of LLaVA-1.5; (ii) Combining the captions from InstructBLIP and LLaVA-1.5, the answer hit rate gets significant improvement and the answer noise rate increases a bit compared with the LLaVA-1.5 as the knowledge source. It makes our GeReA acquires more high-quality captions for the answer prediction and further enhances the performance on OK-VQA dataset; (iii) GeReA remarkably outperforms its corresponding MLLM on almost all categories except for the 'Science and Technology' category when using LLaVA-1.5 as the knowledge source. It indicates the superiority of our generate-reason paradigm in external knowledge acquisition and integration.

---

4. We use the text prompt 'question: {question} the answer: ' to evaluate the zero shot performance in direct answer task on A-OKVQA dataset for InstructBLIP, and use the text prompt 'Question: {} Options: {}. Short answer: ' to evaluate the zero shot performance in multiple choice task. Both tasks use beam search for decoding, where sampling temperature is set to 0. The two text prompts are used for A-OKVQA dateset in original paper. Note that the checkpoints of the used InstructBLIP is not fine-tuned on A-OKVQA dataset due to it is not open source. The checkpoints are available at 'https://storage.googleapis.com/sfr-vision-language-research/LAVIS/models/InstructBLIP/instruct_blip_vicuna7b_trimmed.pth'

5. We use the text prompt '{question}\n answer the question using a single word or phrase.' to evaluate the zero shot performance in direct answer task on A-OKVQA dataset for LLaVA-1.5, and use the text prompt '{question} \n Options: {} \n Answer with the option's letter from the given choices directly.'. Both tasks use beam search for decoding, where sampling temperature is set to 0. The two text prompt are used for A-OKVQA datasets in original paper. The checkpoints of MLLM used in GeReA are available at 'https://huggingface.co/liuhaotian/llava-v1.5-7b'

6. Same as the OK-VQA dataset, we select the first 80 captions in LLaVA-1.5 and the first 40 captions in InstructBLIP for each sample as the knowledge source. Note that the checkpoints of the used InstructBLIP is not fine-tuned on A-OKVQA dataset due to it is not open source.

7. For the blended knowledge source, i.e., InstructBLIP+LLaVA-1.5, we choose the zero shot performance of InstructBLIP as the baseline which is higher than LLaVA-1.5 on A-OKVQA dataset.

8. For each knowledge source, we use the zero shot performance of the corresponding MLLM as the baseline to evaluate the improvement. For the blended knowledge source, i.e., InstructBLIP+LLaVA-1.5, we choose the zero shot performance of LLaVA-1.5 as the baseline which is higher than InstructBLIP on OK-VQA dataset.

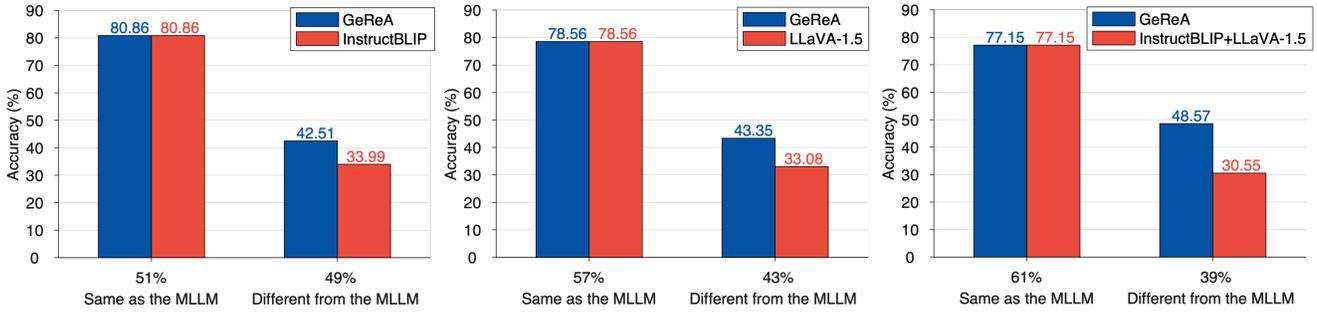

Fig. 4. Two prediction behaviors for GeReA: 'same as the MLLM', and 'different from the MLLM' with three types of knowledge sources.

TABLE 7
The prediction distribution of four situations between GeReA and the MLLM.

| Knowledge Source | MLLM \ GeReA | Correct | Wrong |
|---|---|---|---|
| InstructBLIP | Correct | 50.73% | 11.34% |
|  | Wrong | 15.95% | 21.98% |
| LLaVA-1.5 | Correct | 54.46% | 9.06% |
|  | Wrong | 13.97% | 22.51% |
| InstructBLIP+LLaVA-1.5 | Correct | 56.58% | 6.93% |
|  | Wrong | 14.47% | 22.02% |

### 5.1.2 Prediction Behaviors of GeReA

We define two prediction behaviors for GeReA: 'same as the MLLM', and 'different from the MLLM'. Every testing sample falls into one of the two categories. The statistical results in Fig. 4 show that: (i) For the samples that are same as the corresponding MLLM, the accuracy is over 77%, indicating that they are easy samples for their corresponding knowledge source. (ii) These samples which are different from the MLLM prediction are relatively hard. GeReA has an over 42% accuracy and the corresponding MLLM delivers less than a 33% accuracy. This reveals that GeReA discerningly selects the valuable elements and eliminates the unnecessary ones from the corresponding knowledge source. (iii) With the improvement in the quality of the question-aware prompt captions (InstructBLIP → LLaVA-1.5 → InstructBLIP+LLaVA-1.5), the percentage of testing samples that are same as the corresponding MLLM increases (51% → 57% → 61%), and they achieve high performance (over 77% accuracy). To further explain it, the distribution of four situations between GeReA prediction and the MLLM prediction is displayed in Table 7. From the results, we can see that, as the quality of the question-aware prompt captions improves, the reasoning module of GeReA preserves more correct answers and makes fewer mistakes for the samples where the MLLM prediction is correct. The question-aware prompt caption of higher quality corrects some biases in GeReA. This may be the main reason for the accuracy improvement of GeReA when using different knowledge sources at the same reasoning module. In addition, for the samples where the MLLM prediction is wrong, GeReA provides correct answers for about 40% of them. That is why our GeReA outperforms the MLLM.

In the proposed framework, the MLLM is used to generate question-aware prompt captions. Apart from InstructBLIP and LLaVA-1.5 used in our experiments, other MLLMs that share the same vision encoders (i.e., ViT-g/14 from EVA-CLIP [33] and ViT-L/14 from CLIP [32]) can also be used in our framework, such as InstructBLIP [2], LLaVA [58], LLaVA-1.5 [38], MiniGPT-4 [59], mPLUG-Owl [60], Otter [63], and VPGTrans [81].

### 5.2 Ablation Study

To investigate the effect of each component and their parameters on the performance of GeReA, we conduct extensive ablation studies with three knowledge source settings. Unless specifically noted otherwise, we set the number $N = 5$ for InstructBLIP, LLaVA-1.5, and InstructBLIP+LLaVA-1.5 as our default setting in the below experiments on OK-VQA dataset.

### 5.2.1 Effect of multiple question-specific manual prompts

In this paper, six question-specific prompts with four perspectives for each question relevant image region are conducted by humans. To explore the effectiveness of it, manual prompts are fed into the MLLM in an incremental manner to generate question-aware prompt captions, where the number of manual prompts $n \in {1, 2, 3, 4, 5, 6}$. The results are presented in Fig. 5 and we can find that: (i) With the increase of the number of question-specific prompts, the answer hit rate significantly improves while the answer noise rate notably decreases. GeReA gets more relevant and richer knowledge and the final accuracy grows accordingly.

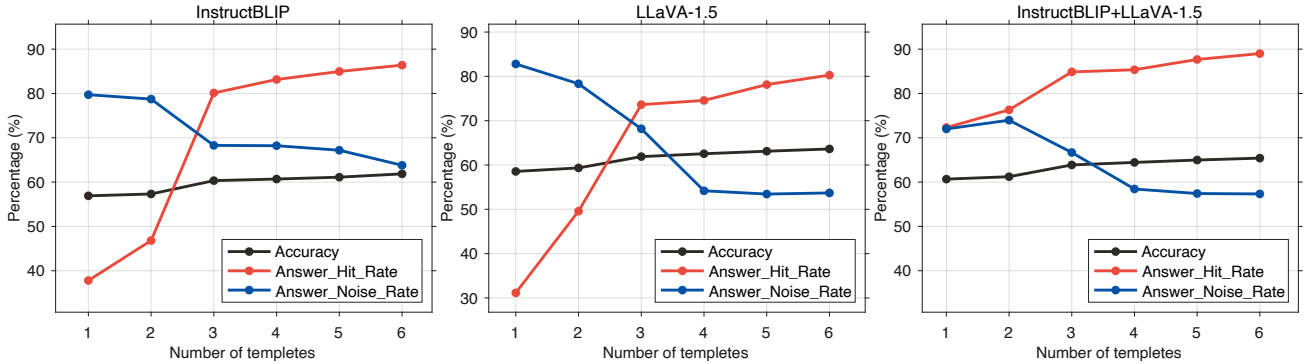

Fig. 5. Performance of GeReA with different number of question-specific manual prompts.

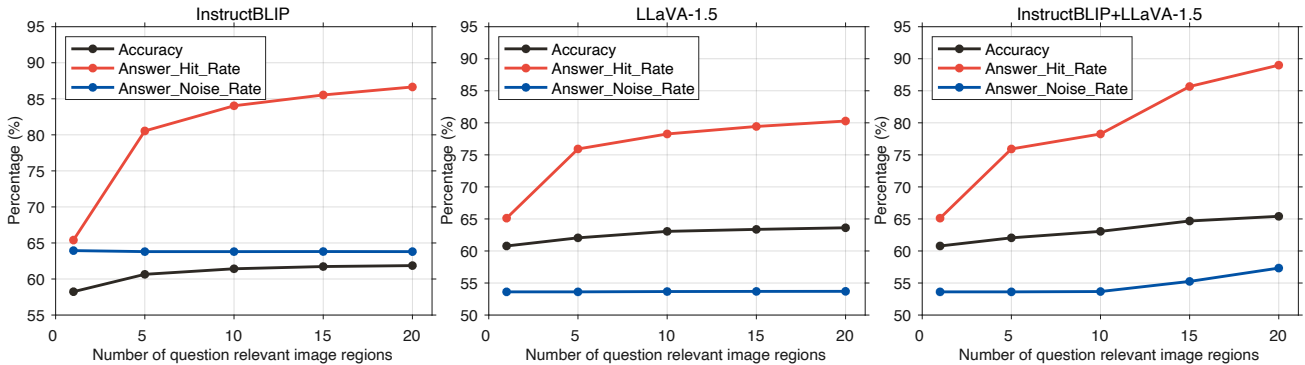

Fig. 6. Performance GeReA with different number of question relevant image regions.

It shows that the question-specific prompt is absolutely crucial for the MLLM to adapt to the knowledge-based VQA task and it can fully activate the capacity of the MLLM as the implicit knowledge engine; (ii) With one prompt, GeReA's accuracy drops by at least 4.96% (number=1 vs. number=6) and it is outperformed by the corresponding MLLM (i.e., InstructBLIP, LLaVA-1.5), showing the importance of multiple question-specific prompts with different perspectives in GeReA.

### 5.2.2 Effect of different numbers of question relevant image regions

Fig. 6 varies the number of question relevant image regions $m$ from 1 to 20 to explore its effect on GeReA. We report the results of accuracy, answer hit rate, and answer noise rate. From the results, we can find that: (i) With the increase of question relevant image regions, the answer hit rate and accuracy grow accordingly but they exhibit a tendency to saturate. This is because the question-relevant visual information is sufficiently acquired from GeReA as the number increases, eventually reaching a saturation point where the quality of question-aware prompt captions stabilizes. (ii) Interestingly, the answer noise rate remains within a stable range and does not vary with the increase of question relevant image regions when using single knowledge source. This can be explained as the fixed interference introduced by the manual question-specific prompt templates for MLLM. It leaves sufficient room for future improvement and implies that the soft question-specific prompt templates may be more suitable for the MLLM to adapt to the knowledge-based VQA task.

### 5.2.3 Effect of similar sample selection

As mentioned above, the similar sample selection strategy is followed by the Prophet [30]. However, other similar sample selection strategies may be better suited for the question-aware prompt caption reasoning stage. To show the effectiveness of different selection strategies, we conduct experiments with three strategies including: (a) fused: our followed strategy that selects similar samples according to the similarity of syncretic features; (b) ques + img: similar samples that are used in PICa [28]; (c) rand: examples that are selected randomly. In Table 8(a), the fused similar sample selection strategy gets the best performance, while the rand similar sample selection strategy yields the the worst performance due to the introduction of irrelevant and noisy information in the three types of knowledge sources. This shows that the similar samples influence the final performance of our GeReA and the fused similar sample selection strategy is more suited for the question-aware prompt caption reasoning stage compared with other similar sample selection strategies.

Besides the similar sample selection strategy, the number of similar samples may also affect the model. Therefore, we conduct experiments with different numbers of the similar samples for the question-aware prompt caption reasoning stage. The numbers are set to 0, 1, 5, and 10, and the results are presented in 8(b). From the results, we can see that: (i) Without any similar sample, GeReA's accuracy drops by at

TABLE 8
Ablation experiments to explore the effect of example selection strategy and different numbers of similar samples for GeReA

(a) Performance of GeReA with different strategies of similar example selection

| Example Selection | Fused [30] | Ques + img [28] | Rand |
|---|---|---|---|
| InstructBLIP | **61.86** | 61.56 | 60.89 |
| LLaVA-1.5 | **63.62** | 63.35 | 62.95 |
| InstructBLIP+LLaVA-1.5 | **65.41** | 65.30 | 64.52 |

(b) Performance of GeReA with different number of similar examples

| Numbers | $N = 0$ | $N = 1$ | $N = 5$ | $N = 10$ |
|---|---|---|---|---|
| InstructBLIP | 61.11 | 61.22 | 61.86 | **62.10** |
| LLaVA-1.5 | 63.17 | 63.23 | **63.62** | 63.50 |
| InstructBLIP+LLaVA-1.5 | 64.75 | 64.87 | **65.41** | 65.17 |

TABLE 9
Performance of GeReA with different caption methods.

| Knowledge Source | Vision Encoder | Dimension | Acc.(%) |
|---|---|---|---|
| InstructBLIP | DETR | (100, 256) | **61.86** |
| | CLIP | (49, 2048) | 61.68 |
| | ResNet | (512, 2048) | 61.43 |
| | w/o Visual Feature | - | 61.37 |
| LLaVA-1.5 | DETR | (100, 256) | **63.62** |
| | CLIP | (49, 2048) | 63.48 |
| | ResNet | (512, 2048) | 63.35 |
| | w/o Visual Feature | - | 63.24 |
| InstructBLIP + LLaVA-1.5 | DETR | (100, 256) | **65.41** |
| | CLIP | (49, 2048) | 65.30 |
| | ResNet | (512, 2048) | 65.17 |
| | w/o Visual Feature | - | 64.96 |

TABLE 10
Performance of GeReA with different caption methods.

| MLLMs | Caption Methods | AHR | ANR | Acc.(%) |
|---|---|---|---|---|
| InstructBLIP | Default | 86.39 | 63.80 | **61.86** |
| | Question-relevant Caption | 58.89 | 85.64 | 57.11 |
| | Generic Caption | 2.67 | 97.33 | 48.45 |
| LLaVA-1.5 | Default | 80.28 | 53.70 | **63.62** |
| | Question-relevant Caption | 60.43 | 84.89 | 57.67 |
| | Generic Caption | 2.98 | 97.02 | 48.48 |

least 0.45% ($N$=0 vs. $N$=5), showing the effectiveness of varying numbers of similar samples in GeReA; (ii) With the increase of the numbers of similar samples, the final accuracy grows accordingly. It indicates that more similar samples provide more useful information for answer prediction. (iii) When $N$=10, the accuracy of GeReA begins to decrease ($N$=5 vs. $N$=10) using LLaVA-1.5 and InstructBLIP+LLaVA-1.5 as knowledge sources. This is because the input length exceeds the max length of T5-large model, resulting in our GeReA losing some useful information for answer prediction.

### 5.2.4 Effect of visual feature

Vision features and different visual encoders may affect the model performance. First, we remove the visual features to explore the effect on GeReA. Then three widely-used vision features are compared below: (i) DETR [71]: detr_resnet101_dc5 which is patch-like feature based on object detection; (ii) CLIP [32]: RN101 which is patch-like feature; (iii) ResNet [82]. For the ResNet, we duplicate the pooled features from ResNet50 to match the length of the text sequence. This is done to simulate patch-like features, where each patch is identical to the pooled image features. The results in Table 9 show that: (i) Without visual feature, our GeReA experiences a decrease compared with its corresponding knowledge source, which indicates the effectiveness of visual feature and is consistent with the findings in REVIVE [25]. (ii) Compared with the corresponding knowledge source, the DETR gets the best performance, while the ResNet gets the worst performance. It indicates that the different vision features affect the model performance in the different knowledge source and the DETR's visual features contain richer visual knowledge.

### 5.2.5 Effect of different caption methods

GeReA achieves the best performance at this moment and the question-aware prompt captions are critical to the proposed framework. In order to further understand the importance of question-aware prompt captions, we compare it with other caption methods used in knowledge-based VQA. The compared caption methods include: (i) Question-relevant Caption [52]: using a BLIP model to generate the captions via the question-relevant portion of the image and a image-text matching model to filter the noisy captions; (ii) Generic Caption [28]: using a VinVL-base captioning model [51] to generate the captions without any question-relevant information from the whole image. These works use the captions as text prompts to activate the capacity of LLM as an implicit knowledge engine for knowledge-based VQA, the knowledge acquisition and integration highly rely on the LLM. Different from these works, our GeReA acquires the relevant knowledge via the frozen MLLM to generate the question-aware prompt captions. The integration of knowledge is achieved by learning a knowledge-image-text representation from the question-aware prompt captions via a multi-modal reasoning model. In the experiments, we replace the question-aware prompt caption with the caption generated by the above methods in our GeReA.

For an equitable comparison, we re-implement the question-relevant caption method and generic caption method with the MLLM which used in our framework. In Table 10, we exhibit the performance of the three caption methods in our GeReA in terms of answer hit rate, answer noise rate, and accuracy respectively. We have the following observations: (i) The answer hit rate of the compared caption methods is less than 60%, while the answer noise rate is more than 85%. The accuracies of the compared caption methods are both lower than the zero shot performance of the corresponding MLLM. It indicates that these methods do not activate the capacity of the MLLM as an implicit knowledge engine which contain vast useless information in these captions. They highly rely on a powerful LLM to jointly acquire and reason over the relevant knowledge. (ii) The generic caption method delivers less than a 3% answer hit rate and achieves an over 48% accuracy in our GeReA. This is because the similar samples and visual feature provide the relevant knowledge for final answer prediction. (iii) Compared to question-relevant caption, our question-aware prompt caption gets an over 80% answer hit rate and less than a 64% answer noise rate. The biggest difference between question-aware prompt caption between question-relevant caption is the question-specific manual prompt. This once again demonstrates that the multiple question-specific prompts with different perspectives are absolutely crucial for the MLLM to adapt to the knowledge-based VQA task and can fully activate the capacity of the MLLM as the implicit knowledge engine.

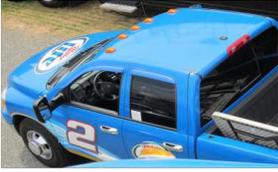

Fig. 7. The samples predicted correctly by GeReA. The question-aware captions successful hit the ground-truth answer and our reasoning model in stage-2 can correctly infer the answer from these captions, even if it includes some irrelevant, erroneous, and misleading information. The ground truth answer is marked in blue and the misleading answer is marked in red.

### 5.3 Case Study

Finally, we provide some success and failure cases along with their question-aware prompt captions to intuitively understand the proposed GeReA in Fig. 7 and Fig. 8.

The samples predicted correctly by GeReA are shown in the Fig. 7. We can see that the question-aware prompt captions successful hit the ground-truth answer and our reasoning model in stage-2 can correctly infer the answer from these captions, even if it includes some irrelevant, erroneous, and misleading information. From the 1st-2nd columns in the 1st row of Fig. 8, we can find that the question-aware prompt captions do not hit the ground-truth answer but our GeReA correctly predict the final answer. This is mainly attributed to the similar samples and visual features introduce relevant knowledge for the answer prediction. From the 3rd-4th columns in the 1st row of Fig. 8, the question-aware prompt captions hit the ground-truth answer, the MLLM correctly predict the answer but our GeReA predict the wrong answer. we obverse that the answer predicted by GeReA is semantically close to the ground truth answer, which is still reasonable under human evaluation. In the last row of Fig. 8, we show the samples incorrectly predicted by our GeReA and the MLLM. From these samples, we draw the following conclusions: (i) The evaluation metric is too rigid to recognize the reasonable answers, i.e., the different expressions of the correct answer. More flexible metric should be developed to evaluate the performance of OK-VQA accurately. (ii) The MLLM might make factual errors, but our GeReA does not correct these errors now. A more powerful reasoning model in stage-2 should be presented in the future. (iii) The question-aware prompt captions do not hit the ground-truth answer for these samples. Blending more knowledge sources may be the solution to this problem.

## 6 CONCLUSION

We propose GeReA, a simple but effective two-stage framework which uses the MLLM as the implicit knowledge engine for knowledge-based VQA. In the first stage, different from the retrieval-based methods and the GPT-3 based methods, GeReA acquires the knowledge by prompting the MLLM with the question-relevant image regions and question-specific manual prompts to generate the question-aware prompt captions. In the second stage, a multi-modal reasoning model is trained to learn a joint knowledge-image-text representation for final prediction. GeReA unlocks the first use of MLLM as the implicit knowledge engine and provides a novel paradigm for knowledge-based VQA. Extensive comparative experiments, ablation experiments and case studies on the two challenging benchmarks, i.e., OK-VQA and A-OKVQA demonstrate that our GeReA achieves the best performance in comparison with all state-of-the-art methods. Notably, the MLLM used in our GeReA is replaceable and the performance of the proposed framework directly depends on the MLLM. This reflecting the scalability and flexibility of our GeReA. We envision our work as a strong benchmark that will catalyze future research in the field of knowledge-based VQA.

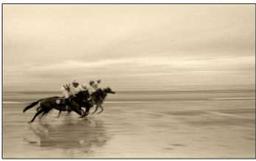

Fig. 8. Four types of different cases with results of different methods. (i) the question-aware prompt captions do not hit the ground-truth answer but our GeReA correctly predict the final answer. (ii) the question-aware captions hit the ground-truth answer, the MLLM correctly predict the answer but our GeReA predict the wrong answer. (iii) the question-aware prompt captions hit the ground-truth answer but our GeReA and the MLLM both predict the wrong answer. (iv) the question-aware prompt captions do not hit the ground-truth answer, our GeReA and the MLLM both predict the wrong answer.

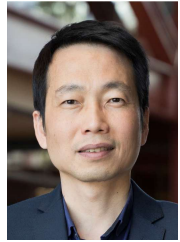

**Jianfei Cai** (S'98-M'02-SM'07-F'21) received his PhD degree from the University of Missouri Columbia. He is currently a Professor and serves as the Head of the Data Science & AI Department at Faculty of IT, Monash University, Australia. Before that, he had served as Head of Visual and Interactive Computing Division and Head of Computer Communications Division in Nanyang Technological University (NTU). His major research interests include computer vision, multimedia and visual computing. He is a co-recipient of paper awards in TPAMI, ACCV, ICCM, IEEE ICIP and MMSP. He serves or has served as an Associate Editor for TPAMI, IJCV, TIP, TMM, and TCSVT as well as serving as Area Chair for CVPR, ICCV, ECCV, IJCAI, ACM Multimedia, ICME and ICIP. He was the Chair of IEEE CAS VSPC-TC during 2016-2018. He has also served as the leading TPC Chair for IEEE ICME 2012 will be the leading general chair for ACM Multimedia 2024. He is a Fellow of IEEE.

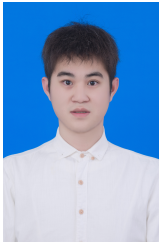

**Ziyu Ma** received the B.S. degree from the Lanzhou University of Technology, Lanzhou, China, in 2019. He is currently pursuing the Ph.D. degree with the Laboratory of Vision and Image Processing, Hunan University, Changsha, China. His research interests include multimodal learning, human–robot natural interaction, multimodal information fusion, and deep learning.

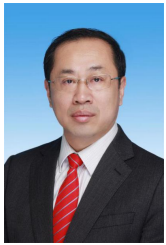

**Shutao Li** (M'07-SM'15-F'19) received the B.S., M.S., and Ph.D. degrees from Hunan University, Changsha, China, in 1995, 1997, and 2001, respectively.In 2001, he joined the College of Electrical and Information Engineering, Hunan University. From May 2001 to October 2001, He was a Research Associate with the Department of Computer Science, Hong Kong University of Science and Technology. From November 2002 to November 2003, he was a Postdoctoral Fellow with the Royal Holloway College, University of London. From April 2005 to June 2005, he was a Visiting Professor with the Department of Computer Science, Hong Kong University of Science and Technology. He is currently a Full Professor with the College of Electrical and Information Engineering, Hunan University. He has authored or co-authored over 200 referred papers. He won the Second Prize National Natural Science Award of China in 2019 and gained two 2nd-Grade State Scientific and Technological Progress Awards of China in 2004 and 2006. His current research interests include image processing, pattern recognition, and artificial intelligence. He is an Associate Editor of the IEEE TRANSACTIONS ON GEOSCIENCE AND REMOTE SENSING and the IEEE TRANSACTIONS ON INSTRUMENTATION AND MEASUREMENT. He is a member of the Editorial Board of the Information Fusion and the Sensing and Imaging.

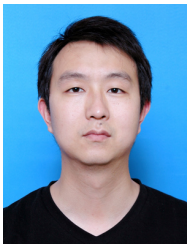

**Bin Sun** (S'15-M'16) received the B.S. and Ph.D. degrees in control science and engineering from Hunan University, Changsha, China, in 2010 and 2016, respectively. From 2017 to 2019, he was a Post-Doctoral Researcher in electrical engineering with the College of Electrical and Information Engineering, Hunan University, where he is currently an Associate Professor. His research interests include computer vision and human-robot natural interaction.

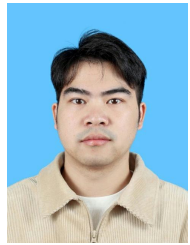

**Zuxiang Long** received the B.S. degree in School of Mechatronic Engineering from China University of Mining and technology, Xuzhou, China, in 2020. He is currently pursuing the master's degree in College of Electrical and Information Engineering, Hunan University. His research interests include image processing, affective computing and model compression.

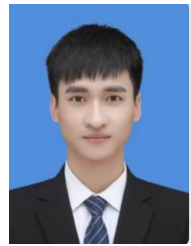

**Fuyan Ma** received the B.S. degree in the College of Electronics and Information Engineering from South Central University for Nationalities, Wuhan, China, in 2018, and the Ph.D. degree in control science and engineering from Hunan University, Changsha, China, in 2023. He is currently a Post-Doctoral Researcher in Defense Innovation Institute, Chinese Academy of Military Science, Beijing, China. His research interests include image processing, affective computing and multimodal information fusion.